\documentclass[runningheads,11pt,a4paper]{llncs}

\usepackage{slashbox}
\usepackage{multirow}
\usepackage{booktabs}
\usepackage{float}
\usepackage{cite}

\usepackage{mathrsfs}
\usepackage{amsmath}
\usepackage{amsfonts}
\usepackage{amssymb}
\usepackage{graphicx}
\usepackage{subfigure}
\usepackage[scriptsize]{caption2}
\usepackage{color}
\usepackage{bm}

\usepackage{geometry}

\usepackage{pdfsync,enumerate}
\usepackage[vlined,ruled,linesnumbered]{algorithm2e}

\usepackage{bm}
\newtheorem{thm}{Theorem}

\newtheorem{cor}{Corollary}
\newtheorem{rem}{Remark}

\def\diag{\mathrm{diag}}

\def\sign{\mathrm{sign}}

\usepackage{amssymb}
\usepackage{graphicx}
\usepackage{graphicx}
\usepackage{epstopdf}
\usepackage{url}
\urldef{\mailsa}\path| yangxinguang@hotmail.com,ylu@vsu.edu|
\urldef{\mailsb}\path||
\urldef{\mailsc}\path||
\newcommand{\keywords}[1]{\par\addvspace\baselineskip
\noindent\keywordname\enspace\ignorespaces#1}

\begin{document}

\mainmatter  

\title{Informative Gene Selection for Microarray Classification via Adaptive Elastic Net with Conditional Mutual Information}

\titlerunning{Adaptive Elastic Net with Conditional Mutual Information}

%
%

\author{Xin-Guang Yang$^{1}$\and Yongjin Lu$^{*}$}
%
\authorrunning{X. Yang, Y. Lu}

\institute{$^{1}$College of Mathematics and Information Science, Henan Normal University,\\
Xinxiang 453007, P. R. China\\
Department of Mathematics and Economics, Virginia State University\\ Petersburg, VA 23806, USA\\
\mailsa\\
\mailsb\\
\mailsc\\
\url{}}

%
%

\toctitle{Lecture Notes in Computer Science11111111111111111111111111}
\tocauthor{Authors' Instructions}
\maketitle

\begin{abstract}
Due to the advantage of achieving a better performance under weak regularization, elastic net has attracted wide attention in statistics, machine learning, bioinformatics, and other fields. In particular, a variation of the elastic net, adaptive elastic net (AEN), integrates the adaptive grouping effect. In this paper, we aim to develop a new algorithm: Adaptive Elastic Net with Conditional Mutual Information (AEN-CMI) that further improves AEN by incorporating conditional mutual information into the gene selection process. We apply this new algorithm to screen significant genes for two kinds of cancers: colon cancer and leukemia. Compared with other algorithms including Support Vector Machine, Classic Elastic Net and Adaptive Elastic Net, the proposed algorithm, AEN-CMI, obtains the best classification performance using the least number of genes.




\keywords{Elastic Net, Conditional mutual information, Gene selection, Microarray data classification.}
\end{abstract}

\section{Introduction}\label{sec1}

A deoxyribonucleic acid (DNA) microarray, a collection of microscopic DNA spots attached to a solid surface, is an important biotechnology that allows scientists to measure the expression levels of a large number of genes simultaneously. One of the main purposes of conducting DNA microarray experiments is to classify biological samples and predict clinical or treatment strategies for certain disease, such as various cancers, using gene expression data. Although a classification problem is not new to statistical or data mining community, gene expression data from DNA microarray experiments exhibits some unique features. Two most important ones are high dimensionality and small sample size, which are due to the fact that the number of genes collected from microarray experiments is much bigger than than the number of available samples \cite{tdp1999,ijsv2002,tjka2013,fywmy2014,ylr2018,zhwgdk2014,cyh2011,jychhz2014,z2015}. Therefore, classifying DNA gene microarray data requires a new set of statistical or data mining methods that could reduce the dimension of the data or select variables of great significance while maintaining a high level of classification accuracy. From the biological perspective, this is equivalent to identifying informative genes associated with the occurrence of the disease under study. There are many results in the literature that address the problem of informative gene selection for DNA microarray classification. A large group of these results \cite{ijs2002,uad2013,ylcjj2013,yme2011,imy2015,yzxyx2014,zyy2015} are based on support vector machine (SVM) and its variants, a class of non-probabilistic machine learning algorithms that seek a nonlinear decision boundary efficiently. For instance, fMRI-based hierarchical SVM was applied to the automatic
classification and grading of liver fibrosis in \cite{yme2011}. A postprocessing SVM \cite{ylcjj2013} was proposed to detect incorrect and correct measurements in real-time continuous glucose
monitoring systems. In \cite{uad2013}, Maulik et al. proposed a novel transductive SVM which achieved a better classification accuracy to select potential gene markers to predict cancer subtypes. Another group of researchers \cite{tibshirani1996,ht2005,djg2010,h2006,hh2009,gn2006,blm2005} address the problem of dimension deduction for microarray gene expression data by adding penalty terms related to the number of features to the cost functional the algorithm aims to minimize. This penalty strategy gives rise to algorithms including: LASSO with $L_1$-norm penalty \cite{tibshirani1996} and its improved variants \cite{ht2005,djg2010,h2006}; sparse logistic or multinomial logistic regression with Bayesian regularization \cite{gn2006, blm2005}.

To improve the classification performance on the DNA microarray gene expression data, the idea of selecting selecting informative genes in groups is exploited due to the reason that complex biological processes, such as, tumor and cancer prediction and diagnosis, are not determined by a single gene but by the interactions of a few genes in groups. The group gene selection counterparts of some of the previously reviewed algorithms are developed. In \cite{my2006}, Yuan et al. presented the group Lasso (GL) algorithm, which is extended to logistic regression by Meier et al. in \cite{lsp2008}. In order to obtain group-wise sparsity and within group sparsity simultaneously, Simon et al. \cite{njt2013} presented a sparse group lasso (SGL) algorithm, which is solved numerically by the method of accelerated generalized gradient descent. The  multi-class classification variant of the multinomial sparse group lasso was then developed in \cite{mn2014}.

Though the sparse group lasso \cite{njt2013} and its improved variants could obtain sparsity within a group by leading into the extra $l_1$-norm penalty, they do not give the biological significance of the genes in the same selected group of genes. To solve the problem of estimating the gene coefficients that represent the importance of individual gene while at the same time performing gene selection, statistical methods such as adaptive LASSO \cite{h2006} and adaptive elastic net \cite{hh2009} that could adaptively select variables should be applied. The adaptive lasso \cite{h2006} is a
weighted $l_1$-penalization method with adaptive weights initially estimated from the data and it enjoys the oracle properties. Subsequently, Zou and Hastie proposed adaptive elastic net, an $l_2$ and $l_1$ mixed penalization method, that could be applied to high-dimension data while maintaining the oracle properties in \cite{hh2009}. This improves the adaptive LASSO and the original elastic net
by combining adaptively weighted $l_1$ penalization term with weights estimated from the original elastic net together with the $l_2$ penalization term from the elastic net. However, when applying the adaptive elastic net to high dimension gene expression data, owing to the requirements of the lower precision, some significant genes might be falsely assigned smaller weight values in the initial estimator. Therefore, these important genes would be incorrectly deleted from the model by shrinkage method, which leads to lower prediction accuracy in informative gene selection for microarray DNA data. In addition, adaptive elastic net might not perform well if the pairwise correlation between variables are not high.

To address these issues, we propose a new method: adaptive elastic net with conditional mutual information (AEN-CMI) that weighs both the $l_1$ and $l_2$ penalization terms with weights estimated based on the \emph{conditional mutual information} among different genes in the data set. How the weights are estimated is a major departure of our method from the well established method of adaptive elastic net. The idea of applying information theory on solving gene selection problems have been explored in the literature (see \cite{sas2011,jlxdx2014} for example) where various information theory based feature selection algorithms have been developed to search the subset of best genes. In particular, conditional mutual information is used to conjecture the gene regulatory networks of cancer gene expression datasets \cite{xxky2012}. By incorporating the conditional mutual information, the conditional dependency among genes, into the adaptive weight estimation, the aforementioned drawbacks of adaptive elastic net could be avoided to certain degree. In this article, we will present a full mathematical description of the new method of AEN-CMI and prove a theorem that explains why our method could encourage a grouping effect. The optimization problem will then be solved by a regularized solution path algorithm -- the pathwise coordinate descent algorithm (PCD). We then evaluate the performance of our algorithm on colon cancer and leukemiaer gene expression datasets. The performance of the proposed algorithm will also be compared with other popular methods, including SVM, classic elastic net and adaptive elastic net. The experiment results show that our algorithm performs the best in the sense that it obtains the highest classification accuracy by using the smallest number of genes.

The rest of the paper will be organized as the following: Section \ref{sec2} briefly states the research problem and the preliminaries. The adaptive elastic net
with conditional mutual information (AEN-CMI) is presented in Section \ref{sec3}. The regularized solution path algorithm, PCD, is developed in Section \ref{sec4}. Experimental results on the two
cancer gene expression datasets are provided in Section \ref{sec5}, and Section \ref{sec6} concludes the whole article.

\section{Problem Statement and Preliminaries}\label{sec2}
Microarray classification in its essence is a binary classification problem, the abstract formulation of which we give below.
For a training set $\{(x_{1},y_{1})$ $,\cdots,(x_{n},y_{n})\}$,
where $x_{i}=(x_{i1},\cdots,x_{ip})$ is the input vector and
$y_i\in\{0,1\}$ denotes its class label, and then the classification
problem is aim to learn a discrimination rule $f: R^{p}\rightarrow
\{0,1\}$. Hence, we can allot a class label to any new sample. For
microarray gene expression data, $n$ and $p$ respectively represent the number of
tumor types and the number of genes. Let $
y=(y_{1},\cdots,y_{n})^{T}$ be the response vector and $X=(
x_{1};\cdots;x_{n})=(x_{(1)},\cdots, x_{(p)})$ be the model matrix.
Let $x_{(j)}=(x_{1j},\cdots,x_{nj})^{T}$ represent the $j$th predictor.
We also assume that the response vector $y$ is centered and the columns of
$X$  are standardized, i.e.,
\begin{eqnarray}
\sum_{i=1}^{n}y_{i}=0,\quad  \sum_{i=1}^{n}x_{ij}=0,\quad
\frac{1}{n}\sum_{i=1}^{n}x_{ij}^{2}=1.
\end{eqnarray}

According to a general linear regression model, we can predict the response vector $y$ by
\begin{eqnarray}
\hat{y}= X\hat{\bm \theta}=\sum_{j=1}^{p}\hat{\theta}_{j} x_{(j)}
\end{eqnarray}
where $\hat{\theta}=(\hat{\theta}_{1},\cdots,\hat{\theta}_{p})^{T}$ is
the estimated coefficient vector. Note that the number of non-zero
estimated coefficients in $\hat{ \theta}$ is equivalent to the
final number of selected genes. Let $I(\hat{y}_{\tau}>0.5)$ indicate
the classification function, where $I(\cdot)$ denotes the indicator
function and $\hat{y}_{\tau}$ is the prediction value for the given
sample $\tau$ by the discrimination rule. Hence, the binary
classification problem could be handled by the models of regression.

The aim of this paper is two-fold: to predict the type of tumor for a new sample and to
automatically select the relevant important genes to a biological process, in particular, the gene selection in microarray data classification. These two challenging problems will be solved using a novel algorithm: Adaptive Elastic Net with Conditional Mutual Information (AEN-CMI) that incorporates conditional mutual information into the variable selection process.

For the sake of completeness, we review some basic definitions from information theory including
entropy, mutual information, conditional mutual information in the next subsection.

\subsection{Information-Theoretic Measures}\label{sub21}
This section states the principles of information theory by
focusing on entropy and mutual information in a concise form. Then we give some
basic concepts about information theory \cite{tj1991}.

Let$\tilde{X}=(\tilde{x}_{1},\cdots,\tilde{x}_{n})$,
$Y=(y_{1},\cdots,y_{n})$ and $Z=(z_{1},\cdots,z_{n})$ be three sets
of discrete random variables. For simplicity, $\log_{2}$ is expressed as simply $\log$ in this
paper. The \textbf{information entropy of variable $\tilde{X}$} can be defined as
$H(\tilde{X})= -\sum_{\tilde{x}\in
\tilde{X}}p(\tilde{x})\log(\tilde{x})$, where $p(\bar{x})$
denotes the probability distribution of each $\bar{x}$. The
entropy of a random variable is an average measure of its
uncertainty.  \textbf{The conditional information entropy of $\tilde{X}$ given $Y$}
is denoted as $H(\tilde{X}|Y)= -\sum_{y\in Y}p(y)\sum_{\tilde{x}\in
\tilde{X}}p(\tilde{x}|y)\log{p(\tilde{x}|y)}$, where
$p(\tilde{x}|y)$ denotes the conditional probability distribution. The
conditional entropy $H(\tilde{X}|Y)$ is the entropy of a
variable $\tilde{X}$ conditional on the given another variable $Y$.

\textbf{Mutual information (MI)} measures the
amount of information shared by $\tilde{X}$ and $Y$ which are used to describe
the degree of correlation between the two variables sets and its definition is as following
\begin{eqnarray}\begin{aligned}\label{eqnb1}
I(\tilde{X};Y)=\sum_{\tilde{x}\in \tilde{X}}\sum_{y\in Y
}p(\tilde{x},y)\log\frac{p(\tilde{x},y)}{p(\tilde{x})p(y)},\end{aligned}\end{eqnarray}
where $p(\tilde{x},y)$ denotes the joint probability of $\tilde{x}$ and
$y$. By the definition of mutual information, the larger $I(\tilde{X};Y)$ is, the more relevant variables $\tilde{X}$ and $Y$ will be.

\textbf{Conditional mutual information (CMI)} measures conditional dependency
between two variables given other variable. The CMI of variables
$\tilde{X}$ and $Y$ given $Z$ is defined as
\begin{eqnarray}\begin{aligned}\label{eqnb3}
I(\tilde{X};Y|Z)= \sum_{\tilde{x}\in \tilde{X}}\sum_{y\in Y
}\sum_{z\in
Z}p(\tilde{x},y,z)\log\frac{p(\tilde{x},y|z)}{p(\tilde{x}|z)p(y|z)},
\end{aligned}\end{eqnarray}
where $I(\tilde{X};Y|Z)$ denotes the amount of information shared
by variables $\tilde{X}$ and $Y$ given variable $Z$. The CMI will become as a particularly important
property in understanding the results of this paper.

\section{Adaptive Elastic Net with Conditional Mutual Information}\label{sec3}
In this section, we propose a strategy of adaptive gene selection, which would be further developed into the AEN-CMI algorithm.

\subsection{Strategy of adaptive  gene selection}\label{sec31}
For cancer gene expression data, variables (genes) $x_{(k)}$ and $x_{(j)}$ are
two vectors, where the elements denote their values of expression in
different conditions or samples. Mutual information usually
describes the degree of correlation between genes $x_{(k)}$ and
$x_{(j)}$. Conditional mutual information not only describes the correlation degree between pairwise genes $\{x_{(k)}, x_{(j)}\}$ given the class label
$y$ but also surveys conditional dependency between two genes
when the class label is given. In the following, we propose a mechanism to assess the importance of $k$th
gene by applying the conditional mutual information.

Define $s_k$ to be the individual significance of  $k$th gene, i.e.,
\begin{eqnarray}\begin{aligned}\label{eqn4}
s_k= \frac{1}{p-1}\sum\limits^{p}_{j=1}I(x_{(k)};x_{(j)}|
y),\end{aligned}
\end{eqnarray}
where $x_{(k)}, x_{(j)}$ respectively denotes the
$k$th, $j$th gene expression level among all the $p$ genes, where
$k,j=1,\cdots,p$. $s_k$ is the class-conditional correlation between the gene
$x_{(k)}$ and all the other genes measures, thus it measures the average information
shared by gene $x_{(k)}$ and remaining genes conditionally on $y$. Here, $s_k$ includes the
complementary information between $x_{(k)}$ and all other genes,
which enables us to assess correlation between genes in groups and could help us make more accurate predictions.  Moreover,
$s_k$ can be used as a quantitative index to measure how significant a gene is, i.e., the higher the value of the $s_k$ is,
the more significant the gene $x_{(k)}$ will be. As an extreme case, $s_k=0$ if the
gene $x_{(k)}$ can not provide useful information for the class
label.

Based on $s_k$ (\ref{eqn4}), we further construct the weight coefficient for the $k$th gene
\begin{eqnarray}\label{eqnw}
\label{eqb1} w_{k}=(s_k+\delta)^{-1},
\end{eqnarray}
where the controllable parameter $0<\delta\ll 1$ is a given
threshold. The $k$th gene has distinct significance when
$s_k> 0$. However, the $k$th gene is not significant in predicting $y$ if $s_k= 0$.
We denote the matrix of weights as
\begin{eqnarray}\begin{aligned}\label{eqn8}
W=\diag(w_{1},\cdots,w_{p}),
\end{aligned}\end{eqnarray} where $k=1,\cdots, p$.
\begin{rem}
The computation of the weights and their meanings are
not given in the multinomial sparse group lasso model \cite{mn2014}.
The initial consistent estimator is used to construct the weights
for the adaptive lasso \cite{h2006} and the initial elastic net
estimator is used to construct the weights for the adaptive
elastic net \cite{hh2009}. Although the above-mentioned two weights
have clear statistical meanings and could be roundly applied to evaluate
the gene importance, they can not indicate the obvious
biological significance. The strategy of adaptive gene selection
presented in this paper has biological significance.
\end{rem}

\subsection{Statistical learning model}\label{sec33}
Utilizing the weight matrix \eqref{eqn8} that contains the conditional mutual information of individual genes, we propose the following penalty term for the adaptive elastic net:
\begin{eqnarray}\label{eqn8b}
(1-\alpha)\|\sqrt{W}\theta\|^2+\alpha\sum_{j=1}^{p}w_{j}|\theta_{j}|,
\end{eqnarray}
where $\|\sqrt{W}\theta\|^2= \sum_{j=1}^{p}w_{j}\theta_{j}^{2}$.

The proposed AEN-CMI algorithm aims to seek the following:
\begin{eqnarray}
\label{eqn8c}\aligned \hat{\theta}(acmi)=\mbox{arg}\min_{\theta}
\|y-X\theta\|^2+\lambda\left((1-\alpha)\|\sqrt{W}\theta\|^2+\alpha\sum_{j=1}^{p}w_{j}|\theta_{j}|\right).
\endaligned
\end{eqnarray}

where $\alpha\in [0,1]$, $\lambda>0$ are the regularization
parameters. Here, we use a squared error loss term.

The connection between the proposed AEN-CMI and some of the classic method is that: the AEN-CMI (\ref{eqn8c}) could be transformed into adaptive
lasso \cite{h2006} if $\alpha=1$; and into elastic net \cite{ht2005} if the weight matrix
$W$ is the identity matrix.

\begin{rem}
In comparison with the adaptive elastic net \cite{hh2009}, the
proposed model \eqref{eqn8c} uses adaptive weights based on
conditional mutual information in instead of ridge regression. Since
the same weight is imposed on both 1-norm penalized coefficient and
2-norm penalized coefficient and conditional mutual information is robust to outliers in dataset, the shrinkage of the adaptive elastic
net with conditional mutual information would produce better
performance in the process of automatic gene selection and have
clear biological significance.
\end{rem}

Since the complex diseases are caused by disruption in gene
pathways rather than individual genes, disease diagnosis using gene expression data should bring insights into the grouping
information. The elastic net algorithms \cite{hh2009,ht2005} are
widely known for encouraging grouping effect. In general, if the regression coefficients of the group with highly correlated variables incline to be equal, then the regression approach can detect the grouping effect. It should be noted that the important genes may
be highly correlated with some inessential genes, the
redundant noise variables could be included in these models. The following theorem shows that the AEN-CMI model
can select the important genes within each group adaptively, which in turn encourages an adaptive grouping effect. The $\tilde{\theta}_j$ and $\tilde{\theta}_l$ in the following theorem are ``significance of gene ranking''.
\begin{thm}
Suppose that the predictors $x_{(j)}$, $j=1, \cdots, p$ are
standardized, if $\hat{\theta}_j(acmi)\hat{\theta}_l(acmi)>0$ holds, then we have
\begin{equation}
\label{eqn9} \aligned |\hat{\theta}_j(acmi)-\hat{\theta}_l(acmi)|\leq
\frac{\|y\|\sqrt{1-\overline{\gamma}
\rho}}{\lambda(1-\alpha)}\sqrt{\tilde{\theta}_j^2+\tilde{\theta}_l^2}
\endaligned
\end{equation}
where $\rho=cor(x_{(j)}, x_{(l)})=x_{(j)}^Tx_{(l)}=\sum_{i=1}^{n}x_{ij}x_{il}$ and $\overline{\gamma}=2|\tilde{\theta}_j\tilde{\theta}_l|/(\tilde{\theta}_j^2+\tilde{\theta}_l^2)$.
\end{thm}

\textbf{Proof} Let
\begin{equation}
\label{eqn10}
L=\|y-X\theta\|^2+\lambda\left((1-\alpha)\|\sqrt{W}\theta\|^2+\alpha\sum_{j=1}^{p}w_{j}|\theta_{j}|\right).
\end{equation}
Note that Equ. (\ref{eqn8}) is an unconstrained convex optimization, the
subgradient of Equ. (\ref{eqn10}) with respect to $\hat{\theta}_j(acmi)$
satisfies
\begin{equation} \label{eqn11}
\left\{\frac{\partial L}{\partial
\theta_k}\right\}_{\theta=\hat{\theta}(acmi)}=0 \hspace{0.5cm} \mbox{if}
\ \ \hat{\theta}_k(acmi)\neq 0.
\end{equation}

\noindent For $j\geq 1$, according to Equ. \eqref{eqn11} we have
\begin{equation}
\label{eqn12} \aligned
-2x_{(j)}^T(y-X\hat{\theta}(acmi))+\lambda\alpha
w_j\mbox{sign}(\hat{\theta}_j(acmi))+2\lambda(1-\alpha) w_j \hat{\theta}_j(acmi)=0.
\endaligned
\end{equation}

\noindent It should be noted that $w_j=\frac{1}{s_k+\delta}>0$. Hence, Equ.
(\ref{eqn12}) can be represented as
\begin{equation}\label{eqn13}
\aligned
 \hat{\theta}_j(acmi)=\frac{1}{\lambda(1-\alpha)}\Big[|\tilde{\theta}_j|x_{(j)}^T(y-X\hat{\theta}(acmi))-\frac{\lambda\alpha}{2} \mbox{sign}(\hat{\theta}_j(acmi))\Big].
\endaligned
\end{equation}

\noindent Similarly to the above, it can be easily obtained that
\begin{equation}\label{eqn14}
\aligned
 \hat{\theta}_l(acmi)=\frac{1}{\lambda(1-\alpha)}\Big[|\tilde{\theta}_l|x_{(l)}^T(y-X\hat{\theta}(acmi))-\frac{\lambda\alpha}{2} \mbox{sign}(\hat{\theta}_l(acmi))\Big].
\endaligned
\end{equation}

\noindent Note that $\hat{\theta}_j(acmi)\hat{\theta}_l(acmi)>0$, i.e.,
$\mbox{sign}(\hat{\theta}_j(acmi))=\mbox{sign}(\hat{\theta}_l(acmi)).$
We subtract Equ. (\ref{eqn13}) from Equ. (\ref{eqn14})
\begin{equation}\label{eqn15}
\aligned
 \hat{\theta}_j(acmi)-\hat{\theta}_l(acmi)=\frac{1}{\lambda(1-\alpha)}\Big[(|\tilde{\theta}_j|
x_{(j)}-|\tilde{\theta}_l| x_{(l)})^{T}(y-X\hat{\theta}(acmi))\Big]
\endaligned
\end{equation}
According to Equs. \eqref{eqn8} and \eqref{eqn10}, we can obtain
\begin{equation*}
\|y-X\hat{\theta}(acmi)\|^2\leq {L}(\hat{\theta}(acmi))\leq
L(\theta=0)=\|y\|^2.
\end{equation*}
Hence $\label{eqn16} \|(y-X\hat{\theta}(acmi))\|\leq \|y\|.$

\noindent From Equ. (\ref{eqn9}), it can be easily obtained
\begin{equation}
\label{eqn17} \aligned &\|\tilde{\theta}_j x_{(j)}-\tilde{\theta}_l
x_{(l)}\|\\
=&\sqrt{\sum_{i=1}^{n}x_{ij}^2\tilde{\theta}_j^{2}+\sum_{i=1}^{n}x_{il}^2\tilde{\theta}_l^{2}-2|\tilde{\theta}_j\tilde{\theta}_l|x_{(j)}^Tx_{(l)}}\\
\leq&\sqrt{\tilde{\theta}_j^{2}+\tilde{\theta}_l^{2}-2|\tilde{\theta}_j\tilde{\theta}_l|x_{(j)}^Tx_{(l)}}\\
=&\sqrt{\tilde{\theta}_j^{2}+\tilde{\theta}_l^{2}}\sqrt{1-\overline{\gamma}\rho}.
\endaligned
\end{equation}

\noindent Finally, we substitute Equ. \eqref{eqn16} and Equ. \eqref{eqn17} and
Equ. \eqref{eqn15} yields \eqref{eqn9} which completes the proof.

\hfill$\square$

Note that Theorem 1 still holds for the case $|\tilde{\theta}_j|\geq \eta$ and $|\tilde{\theta}_l|<\eta$. The only difference of representation
is substituting $|\eta|$ for $\tilde{\theta}_l$. If the $|\tilde{\theta}_j|\leq \eta$ and $|\tilde{\theta}_l|\leq\eta$, then the following Corollary can be easily obtained.

\begin{cor}
Assume that the predictors $x_{(j)}, j=1,\cdots,p$ are standardized. Let $(\hat{\theta}_{0}(acmi),\hat{\theta}(acmi))$ denote the optimal solution of Equ. \eqref{eqn8c}. If $|\tilde{\theta}_j|\leq \eta$, $|\tilde{\theta}_l|\leq\eta$ and $\hat{\theta}_j(acmi)\hat{\theta}_l(acmi)>0$, then
\begin{equation}
\label{eqn18}
\aligned |\hat{\theta}_j(acmi)-\hat{\theta}_l(acmi)|\leq
\frac{\|y\|\times\eta}{\lambda(1-\alpha)}\sqrt{\tilde{\theta}_j^2+\tilde{\theta}_l^2}.
\endaligned
\end{equation}
\end{cor}

\begin{rem}
It should be noted that the $\eta$, the parameter that quantitatively describes the grouping
effect, in Equ. \eqref{eqn18} is far less than 1. Therefore, AEN-CMI has stronger grouping effect for the case $|\tilde{\theta}_j|\leq\eta\ll 1$, $|\tilde{\theta}_l|\leq\eta\ll 1$ in comparison with Elastic Net.
This implies that more genes are deleted together by $l_{1}$-norm shrinkage if
they are less important to the classification. On the basis of Theorem 1, AEN-CMI can allot identical coefficients to the genes only
if $\rho=1$ and $|\hat{\theta}_j(acmi)|=|\hat{\theta}_l(acmi)|>\eta$. It is shown that
given two gene groups with similar significance of the ranking ($|\tilde{\theta}_j|=|\tilde{\theta}_l|$), the one with more genes should have a larger group size.
This implies that
AEN-CMI can adaptively control the size of the selected
groups and thus adaptively select the important
genes within each group by assessing the significance of the gene ranking.
\end{rem}
\section{Algorithm}\label{sec4}
In this section, we give the algorithm for solving the Adaptive
Elastic Net with Conditional Mutual Information (AEN-CMI) on the
augmented space, and hence some popular algorithms, such as LASSO,
Forward Stagewise and LARS \cite{zyy2015,tibshirani1996}, can be used to solve these models
efficiently.
\begin{algorithm}[!ht]
\caption{Solving Algorithm for AEN-CMI}\label{Aglorc}
\KwIn{The matrix $X$, class label $y$ and the controllable parameter $\delta=0.001$ and regularization parameter
$\alpha=0.05$. }
\KwOut{The gene set $\mathcal{G}$ corresponding to the non-zero coefficients of $\hat{\theta}(acmi)$ .}
      $\mathcal{G}\leftarrow \emptyset$\;
 \For{$k=1$ \KwTo $p$}
        {
        $s_k\leftarrow\frac{1}{p-1}\sum\limits^{p}_{j=1}I(x_{(k)};x_{(j)}|y)$\;
        $w_{k}\leftarrow(s_k+\delta)^{-1}$\;
        }
        $W\leftarrow \diag(w_{1},\cdots,w_{p})$\;
        Solve the adaptive elastic net with conditional mutual information model\;

Let ~$X$ and ~$y$ be the input and output matrix respectively\;
Determine the regularization parameter $\lambda$ by cross-validation\;
Solve the elastic net with penalty factor by using PCD algorithm. Suppose we have estimates $\tilde{\theta}_l$ for $l\neq j$, and we wish to partially optimize with respect to $\theta_{j}$. Based on \cite{jth2007}, the coordinate-wise update has the form:
        $\theta_{j}\leftarrow \frac{S(\frac{1}{n}\sum_{i=1}^{n}x_{ij}(y_{j}-\tilde{y}_{i}^{(j)}),\lambda\alpha)}{1+\lambda(1-\alpha)}$
        where $\tilde{y}_{i}^{(j)}=\tilde{\theta}_{0}+\sum_{l\neq
        j}x_{il}\tilde{\theta}_{l}$ is the fitted value excluding the
        contribution from $x_{ij}$. Let $S(z,\gamma)$ be the
        soft-thresholding operator which is
        \begin{eqnarray*}
        \small \hspace{-1.6cm}\label{eqb2}
            (S(z,\gamma))_{j}=\sign(z_{j})(|z_{j}|-\gamma_{j})_{+}=
           \begin{cases}
           z_{j}-\gamma_{j}, &\mbox{if
        $z_{j}> 0 , \gamma_{j}< |z_{j}|$}\\
        z_{j}+\gamma_{j}, &\mbox{if
        $z_{j}< 0 , \gamma_{j}< |z_{j}|$}\\
            0, &\mbox{if $\gamma_{j} \geq |z_{j}|$}.
           \end{cases}
          \end{eqnarray*}
        Obtain the optimal coefficients $\hat{\theta}(acmi)$\;
Construct the classifier$f(x)=\sign(\hat{\theta}(acmi)x+\hat{\theta}_0)$, and predict the new input data $x$\;
Extract the non-zero coefficients and output the corresponding gene set $\mathcal{G}$\;
\Return $\mathcal{G}$.
\end{algorithm}
It should be noted that there are $p\gg n$ observations and $p$
predictors on the augmented space and $p$ is very large for cancer
microarray data. Thus, solving the optimization problem proposed in AEN-CMI numerically is computational expensive. To this end, we select the method of pathwise coordinate descent algorithm (PCD) due to its fast computational speed for large $p$, $n$ and its availability in the solving package ``glmnet'' of R.


This section gives the procedure of the an efficient solving Algorithm for AEN-CMI to select the optimal gene subset with informative genes, which is now detailed in Algorithm \ref{Aglorc}.

\section{Experiment Results}\label{sec5}

\subsection{Colon Cancer Dataset}\label{sec5.1}
To test the effectiveness of the adaptive elastic net with
conditional mutual information (AEN-CMI), we conduct experiments on
the colon cancer gene expression data. The aim of the colon data
\cite{ijsv2002,cyh2011} is to distinguish the cancerous tissues from
the normal colon tissues.  This data is available online:
\url{http://www.weizmann.ac.il/mcb/UriAlon/download/downloadable-data}.
The colon data is obtained from 22 normal and 40 colon cancer tissues. Gene
expression information is extracted from DNA microarray data
resulting, after pre-processing, in a matrix containing the
expression of the 2,000 genes with highest minimal intensity across
the 62 tissues. Since there is no defined training and test set, we
split the data randomly into a training set of 31 samples and a testing set of the other 31 samples.

\begin{figure}[!htb]
  \centering
  \includegraphics[width=0.50\textwidth]{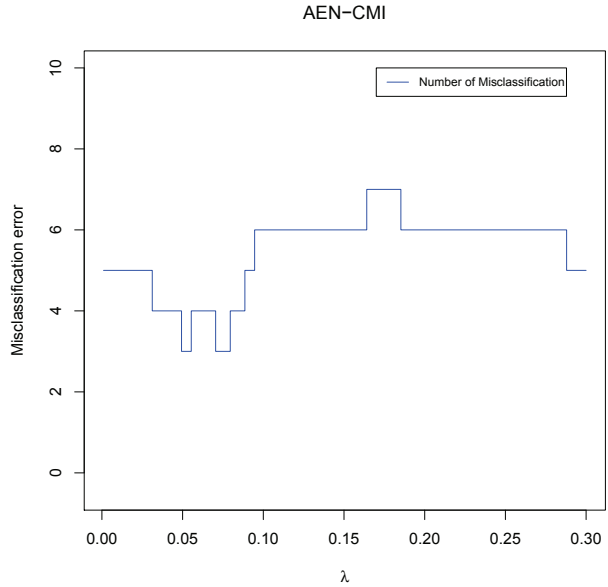}
  \caption{The prediction error of the adaptive elastic net with conditional mutual information}
  \label{fig1}
\end{figure}

Then, we compute the adaptive weight matrix $W$ by equations \eqref{eqnw} and \eqref{eqn8}. Finally, we
solve AEN-CMI with the penalty factor $W$ and determine the relevant
genes. The curve of misclassification errors for the initial elastic
net with the penalty factor $W$ are displayed in Fig. \ref{fig1}. It
is shown that the number of misclassified genes decreases as
the number of the adaptive adjustment increases. However,
it should be noted that the classification accuracy is not
improved when the number of the adaptive adjustment becomes larger. In
fact, using the third adaptive adjustment results in worse classification accuracy. It is also shown that the minimum of
misclassification error (about 3) is obtained when $\lambda\in
[0.05,0.1]$.

\begin{figure}[!htb]
  \centering
  \includegraphics[width=0.50\textwidth]{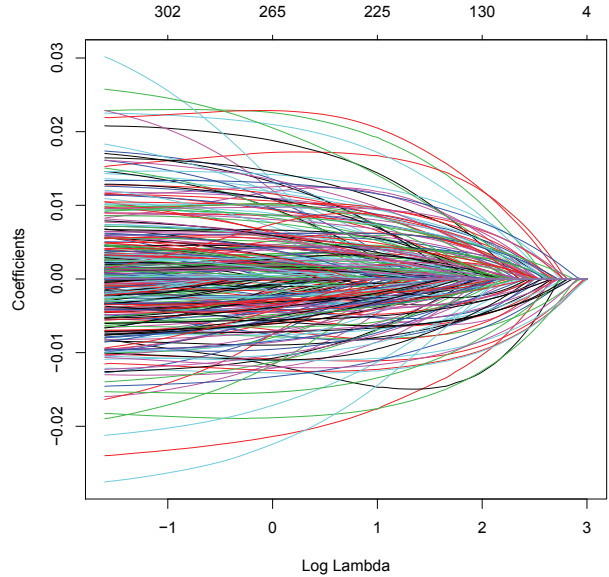}
  \caption{The solution paths of the adaptive elastic net with conditional mutual information (see online version
for colors)}
  \label{fig2}
\end{figure}
Next, we focus on the solution paths of AEN-CMI. To this end, we
randomly spilt the data set into two parts: two-thirds for training
and one-third for testing. The solution paths for the proposed algorithm AEN-CMI is illustrated in
Fig. \ref{fig2}. The horizontal axis represents the natural
logarithm of the parameter $\lambda(\log(X))$, the vertical axis
represents the values of coefficient and each line corresponds to a
coefficient path for a particular gene. Note that any
line segment between two inflection points is linear. Hence, every
coefficient path of AEN-CMI is piecewise linear with respect to
$\log(\lambda)$.

We compare AEN-CMI with the $L_{1}$-SVM, Elastic Net, and Adaptive
Elastic Net (AEN) based on two measurements: classification accuracy and
gene selection performance. The entire process is repeated 10 times
and the results are summarized in Table \ref{table1}. As shown in
the second column of the Table \ref{table1}, AEN-CMI achieves the best
average classification accuracy: higher than AEN and much higher than the other
two methods. The standard deviations for AEN-CMI is the least among all the methods. This implies that AEN-CMI is stabler than the other methods.

As shown in the third column of the Table \ref{table1},  AEN-CMI and
AEN select similar average number of genes which is much less
than that for the other two methods.  The  adaptive strategy
contributes  to the improved properties of gene selection for
the both methods. AEN-CMI achieves the least standard deviations for the number of selected genes
among the four methods.


\begin{table}[!htb]
    \begin{center}
    \caption{Experimental results on the colon cancer gene expression data over 10 runs (the standard deviations are reported in parentheses).}
    \label{table1}
    \begin{tabular}{lll}\toprule
    \multicolumn{1}{l}{\multirow{2}{6cm}{Method}}&\multicolumn{1}{c}{\multirow{2}{5.2cm}{Average classification accuracy}}&\multicolumn{1}{c}{\multirow{2}{5.2cm}{Average number of selected genes}}
    \\  \\ \midrule
   $L_{1}$-svm                      &0.7651 (0.049)        & 52.11 (4.73)   \\
   Elastic Net                      &0.7803 (0.032)        & 67.54 (4.51)   \\
   AEN                              &0.8432 (0.042)        & 25.21 (3.31) \\
   AEN-CMI                          &0.8512 (0.012)        & 24.43 (1.52) \\ \bottomrule
    \end{tabular}
    \end{center}
    \end{table}

     \begin{table*}[!htb]
    \begin{center}
    \caption{Some key genes selected by AEN-CMI over 10 runs on the colon cancer gene expression data}
    \label{table2}
  \begin{tabular}{*{1}{p{2.2cm}}p{2cm}p{9.5cm}cp{2.2cm}}\toprule
  EST name        & GenBank Acc No    &Gene description &  Selected frequency\\[6pt]\hline\\
  Hsa.8147        & M63391            &Human desmin gene, complete cds.&\  \ \ 9/10 \\
  Hsa.36689       & Z50753            &H.sapiens mRNA for GCAP-II/uroguanylin precursor.&\ \ \  10/10\\
  Hsa.3152        & D31885            &Human mRNA (KIAA0069) for ORF (novel proetin), partial cds.&\  \ \ \ 9/10 \\
  Hsa.42186       & AVPR1A            &positive regulation of cellular pH reduction&\  \ \ 7/10  \\
  Hsa.2487        & D14812            &Human mRNA for ORF, complete cds.&\  \ \ 10/10 \\
  Hsa.3306        & X12671            &Human gene for heterogeneous nuclear ribonucleoprotein (hnRNP) core protein A1&\  \ \ \ 8/10 \\
  Hsa.1920        & X06614            &Human mRNA for receptor of retinoic acid.&\  \ \ 10/10 \\
  Hsa.692         & M76378            &Human cysteine-rich protein (CRP) gene, exons 5 and 6.&\  \ \ 10/10 \\
  Hsa.447         & U06698            &Human neuronal kinesin heavy chain mRNA, complete cds.&\  \ \ 9/10 \\
  Hsa.1588        & U09587            &Human glycyl-tRNA synthetase mRNA, complete cds.&\  \ \ 10/10 \\
  Hsa.2051        & X01060            &Human mRNA for transferrin receptor.&\  \ \ 8/10 \\
  Hsa.41260       & L11706            &Human hormone-sensitive lipase (LIPE) gene, complete cds.&\  \ \ 7/10 \\\bottomrule
\end{tabular}
    \end{center}
    \end{table*}

Table \ref{table2} lists the top twelve genes and the selected
frequency in experiments. Compared with AEN, AEN-CMI achieves almost
the same test error, while it selects less genes and achieves least
standard deviations.  It should be also noted that AEN-CMI can
achieve adaptive grouping effect in gene selection. For example,
Hsa.3306 and Hsa.692 are selected in the same group. From the
computational point of view, the regularization parameter and the
kernel parameter of $L_{1}$-svm is optimized by searching a two
dimensional grid of different values for both parameters, which can
slow down the computation considerably. The regularization parameters can
be optimized by the regularization solution path algorithm (PCD) in
AEN-CMI.

\subsection{Leukemia Cancer Dataset}\label{sec5.2}
To illustrate the effectiveness of AEN-CMI, we also conduct
experiments on leukemia dataset \cite{ijsv2002} which includes the
expression profiles of 7129 genes in 47 acute lymphoblastic leukemia
(ALL) and 25 acute myeloid leukemia (AML). This data is available
on-line: \url{http://portals.broadinstitute.org/cgi-bin/cancer/publ
ications/pub_paper.cgi?mode=view\&paper_id=43}. This data set is
preprocessed  as in \cite{ijsv2002}. After preprocessing,
3571 most significant genes are selected for module detection. We
let the label of 47 ALL samples be 0 and 25 AML samples be 1. In the
following, we split the data randomly into 43 training data and 29
test data for the two types of acute leukemia.
\begin{table}[!htb]
    \begin{center}
    \caption{Experimental results on the leukemia cancer gene expression data over 10 runs (the standard deviations are reported in parentheses).}
    \label{table3}
    \begin{tabular}{lll}\toprule
    \multicolumn{1}{l}{\multirow{2}{6cm}{Method}}&\multicolumn{1}{c}{\multirow{2}{5.2cm}{Average classification accuracy}}&\multicolumn{1}{c}{\multirow{2}{5.2cm}{Average number of selected genes}}
    \\  \\ \midrule
   $L_{1}$-svm                      &0.8002 (0.065)        & 54.32 (5.33)   \\
   Elastic Net                      &0.7983 (0.043)        & 46.54 (5.03)   \\
   AEN                              &0.8211 (0.031)        & 28.00 (2.02) \\
   AEN-CMI                          &0.8398 (0.020)        & 23.43 (1.69) \\ \bottomrule
    \end{tabular}
    \end{center}
    \end{table}

We also compare AEN-CMI with the $L_{1}$-SVM, Elastic Net, and
Adaptive Elastic Net (AEN) on leukaemia cancer dataset based on classification accuracy and gene selection performance. The
entire process is repeated 10 times and the results are summarized
in Table \ref{table3}. As shown in the Table
\ref{table3}, AEN-CMI achieves the best average classification
accuracy with the least standard deviation and the least average number of genes selected again with the smallest standard deviation.

     \begin{table*}[!htb]
    \begin{center}
    \caption{Some key genes selected by AEN-CMI over 10 runs on the colon cancer gene expression data}
    \label{table4}
    \small
  \begin{tabular}{*{1}{p{2.5cm}}p{2cm}p{9.2cm}cp{2.2cm}}\toprule
  EST name        & GenBank Acc No   &Gene description &  Selected frequency\\[6pt]\hline\\
  M27891$\_$ at         & CST3              &Cystatin C (amyloid angiopathy and cerebral hemorrhage)     &\  \ \ 10/10 \\
  M84526$\_$ at         & DF                  &D component of complement (adipsin)&\  \ \ \ 10/10 \\
  U18271$\_$ cds1$\_$ at   & MPO               &Myeloperoxidase                                             &\ \ \  8/10\\
  D87024$\_$ at         & GB                  &DEF = (lambda) DNA for immunoglobin light chain&\  \ \ 9/10  \\
  M31166$\_$ at         & PTX3                &Pentaxin-related gene, rapidly induced by IL-1 beta&\  \ \ 10/10 \\
  X57809$\_$ at         & IGL                 &Immunoglobulin lambda light chain&\  \ \ \ 9/10 \\
  M13792$\_$ at      & ADA                 &Adenosine deaminase  1&\  \ \ 10/10 \\
  M98399$\_$s$\_$ at    & CD36                &CD36 antigen (collagen type I receptor, thrombospondin receptor)&\  \ \ 9/10 \\
  U77948$\_$ at         & KAI1                &Kangai 1 (suppression of tumorigenicity 6, prostate; CD82 antigen (R2 leukocyte antigen, antigen detected by monoclonal and antibody IA4))&\  \ \ 8/10 \\
  U05572$\_$ s$\_$ at   & MANB                & Mannosidase alpha-B (lysosomal)&\  \ \ 9/10 \\\bottomrule
\end{tabular}
    \end{center}
    \end{table*}

The summaries of the 10 top-ranked informative genes found by
AEN-CMI for leukaemia cancer gene expression dataset are shown in
Table \ref{table4}, we give the comparisons for both group case and
ungroup case. The important genes obtained by AEN-CMI are emphasized
with bold. Some of these genes
included in the frequently selected gene sets are biologically verified to be
mostly and functionally related to carcinogenesis or tumor histogenesis: For
example, in Table 3, the most frequently selected gene set of
AEN-CMI, including cystatin C (CST3) and myeloperoxidase (MPO) genes
are experimentally proved to be correlated to leukemia of ALL or
AML. The cystatin C gene is located at the extracellular region of
the cell and has role in invasiveness of human glioblastoma cells.
Decrease of cystatin C in the CSF might contribute to the process of
metastasis and spread of the cancer cells in the leptomeningeal
tissues. Matsuo et al. \cite{matsuo2003} believed that the percentage of MPO-positive
blast cells is the most simple and useful factor to predict a
prognosis of AML patients in this category. Other examples are
genes CST3, MPO and IGL highly correlated with the occurrence of
leukaemia. They are selected by AEN-CMI in
the same group owing to the adaptive grouping effect.

\section{Conclusion}\label{sec6}
We propose a novel algorithm: adaptive elastic net with conditional mutual information in this paper.
It is shown that the proposed learning algorithm encourages an adaptive grouping effect and reduces the
influence of the wrong initial estimation to gene selection and
microarray classification. It is also shown that the AEN-CMI
encourages an adaptive grouping effect by evaluating the gene ranking
significance. A fast-solving algorithm is also developed to implement
the proposed AEN-CMI method numerically. Applications of AEN-CMI to two
cancer microarray data sets show that the proposed AEN-CMI algorithm outperforms other algorithms, such as $L_1$-SVM, Elastic Net and classic Adaptive Elastic Net algorithms.


\subsubsection*{Acknowledgments.} Xinguang Yang and Yongjin Lu were supported by the Key Project
of Science and Technology of Henan Province (Grant No. 182102410069).

%
%
%
%
%

\begin{thebibliography}{4}
\bibitem{tdp1999}
T. Golub, D. Slonim, P. Tamayo, et al. ``Molecular classification of
cancer: class discovery and class prediction by gene expression
monitoring,'' \emph{Science}, vol, 286, no, 5439, pp. 531-536, 1999.

\bibitem{ijsv2002}
I. Guyon, J. Weston, S. Barnhill, and V. Vapnik,  ``Gene selection
for cancer classification using support vector machine,"
\emph{Machine Learning}, vol. 46, no. 1, pp. 389-422, 2002.


\bibitem{tjka2013}
T. Amaral, S. J. McKenna,  K. Robertson, and A. Thompson,
``Classification and immunohistochemical scoring of breast tissue
microarray spots,'' \emph{IEEE Transactions on Biomedical
Engineering}, vol, 60, no, 10, pp. 2806-2814, 2013.

\bibitem{fywmy2014}
F. Zhang, Y. Song, W. D. Cai, M. Z. Lee, Y. Zhou, et al, ``Lung
nodule classification with multilevel patch-based context
analysis,'' \emph{IEEE Transactions on Biomedical Engineering}, vol,
61, no, 4, pp. 1155-1166, 2014.


\bibitem{ylr2018}
Y. Wang, X. Li, and R. Ruiz, ``Weighted general group lasso for gene selection in cancer classification"
\emph{IEEE Transactions on Cybernetics}, doi:10.1109/TCYB.2018.2829811, 2018.

\bibitem{zhwgdk2014}
Z. Sun, H. Wang, W. Lau, G. Seet, D. Wang, and K. Lam. ``Microarray
data classification using the spectral-feature-based TLS ensemble
algorithm,'' \emph{IEEE Transactions on NanoBioscience}, vol, 13,
no, 3, pp. 289-299, 2014.

\bibitem{jychhz2014}
J. X. Liu, Y. Xu, C. H. Zheng, H. Kong, and Z. H. Lai, ``Rpcabased
tumor classification using gene expression data,"  \emph{IEEE/ACM
Transactions on Computational Biology and Bioinformatics}, vol. 12,
no. 4, pp. 1-1, 2014.

\bibitem{cyh2011}
C. Zheng, Y. Chong, H. Wang, ``Gene selection using independent
variable group analysis for tumor classification". \emph{Neural
Computing and Applications},  vol. 20, pp. 161¨C170, 2011.

\bibitem{z2015}
Z. Yu, L. Li, J. Liu, and G. Han, ``Hybrid adaptive classifier
ensemble," \emph{IEEE Transactions on Cybernetics}, vol. 45, no. 2,
pp. 177-190, 2015.

\bibitem{ijs2002}
I. Guyon, J. Weston, S. Barnhill, and V. Vapnik, ``Gene selection
for cancer classification using support vector machine,''
\emph{Machine Learning}, vol, 46, no, 1-3, pp. 389-422, 2002.


\bibitem{yme2011}
Y. Sela,  M. Freiman, E. Dery, et al, ``fMRI-based hierarchical SVM
model for the classification and grading of liver fibrosis,''
\emph{IEEE Transactions on Biomedical Engineering}, vol. 58, no. 9,
pp. 2574-2581, 2011.


\bibitem{ylcjj2013}
Y. Leal, L. Gonzalez-Abril, C Lorencio, J. Bondia, and J. Vehi,
``Detection of correct and incorrect measurements in real-time
continuous glucose monitoring systems by applying a postprocessing
support vector machine,'' \emph{IEEE Transactions on Biomedical
Engineering}, vol. 60, no. 7, pp. 1891-1899, 2013.


\bibitem{uad2013}
U. Maulik, A. Mukhopadhyay, and D. Chakraborty,
``Gene-expression-based cancer subtypes prediction through feature
selection and transductive SVM,'' \emph{IEEE Transactions on
Biomedical Engineering}, vol. 60, no. 4, pp. 1111-1117, 2013.

\bibitem{imy2015}
I. Sen,  M. Saraclar,  Y. P. Kahya, ``A comparison of SVM and
GMM-based classifier configurations for diagnostic classification of
pulmonary sounds,'' \emph{IEEE Transactions on Biomedical
Engineering}, vol. 62, no. 7, pp. 1768-1776, 2015.


\bibitem{yzxyx2014}
Y. Tian, Z. Qi, X. Ju, Y. Shi, and X. Liu, ``Nonparallel support
vector machines for pattern classification,'' \emph{IEEE
Transactions on Cybernetics}, vol. 44, no, 7, pp. 1067-1079, 2014.


\bibitem{zyy2015}
Z. Qi, Y. Tian, and Y. Shi, `` Successive overrelaxation for
laplacian support vector machine,'' \emph{IEEE Transactions on
Neural Networks and Learning Systems}, vol. 26, no, 4, pp. 674-683,
2015.

\bibitem{tibshirani1996}
R. Tibshirani,  ``Regression shrinkage and selection via the
lasso,'' \emph{Journal of the Royal Statistical Society, Series B
(Methodological)}, vol. 58, no. 1, pp. 267-288, 1996.



\bibitem{ht2005}
H. Zou and T. Hastie, ``Regularization and variable selection via
the elastic net,'' \emph{Journal of the Royal Statistical Society
B}, vol. 67, no. 2, pp. 301-320, 2005.


\bibitem{djg2010}
D. Angelosante, J. A. Bazerque, and G. B. Giannakis,  ``Online
adaptive estimation of sparse signals: where rls meets the
`1-norm,'' \emph{IEEE Transactions on Signal Processing},vol. 58,
no. 7, pp. 3436-3447, 2010.


\bibitem{h2006}
H. Zou, ``The adaptive lasso and its oracle properties,''
\emph{Journal of the American Statistical Association}, vol. 101,
no. 476, pp. 1418-1429, 2006.

\bibitem{hh2009}
H. Zou and H. H. Zhang, ``On the adaptive elastic net with a
diverging number of parameters". \emph{Annals of Statistics}, vol.
37, no. 4, pp. 1933¨C1751, 2009.

\bibitem{gn2006}
G. C. Cawley and N. L. C. Talbot, ``Gene selection in cancer
classification using sparse logistic regression with Bayesian
regularisation,'' \emph{Bioinformatics}, vol. 22, no. 19, pp.
2348-2355, 2006.


\bibitem{blm2005}
B. Krishnapuram, L. Carin, M. A. Figueiredo, and A. J. Hartemink,
``Sparse multinomial logistic regression: fast algorithms and
generalization bounds,'' \emph{IEEE Transactions on Pattern Analysis
and Machine Intelligence}, vol. 27, no. 6, pp. 957-968, 2005.


\bibitem{my2006}
M. Yuan and Y. Lin, ``Model selection and estimation in regression
with grouped variables''. \emph{Journal of the Royal Statistical
Society, Series B}, vol. 68, no. 1, pp. 49-67, 2006.

\bibitem{lsp2008}
L. Meier, S. van de Geer, and P. B$\ddot{u}$hlmann,  ``The group
lasso for logistic regression,'' \emph{Journal of the Royal
Statistical Society Series B}, vol. 70, pp. 53-71, 2008.


\bibitem{njt2013}
N. Simon, J. Friedman, T. Hastie, and R. Tibshirani,  ``A
sparse-group lasso,'' \emph{Journal of Computational and Graphical
Statistics}, vol. 22, no. 2, pp. 231-245, 2013.

\bibitem{mn2014}
M. Vincent, N. R. Hansen, ``Sparse group lasso and high dimensional
multinomial classification,'' \emph{Computational Statistics and
Data Analysis}, vol. 71, pp. 771-786, 2014.

\bibitem{sas2011}
S. Ghorai, A. Mukherjee, S. Sengupta, and P. K. Dutta,``Cancer
classification from gene expression data by nppc ensemble,"
\emph{IEEE/ACM Transactions on Computational Biology and
Bioinformatics}, vol. 8, no. 3, pp. 659-671, 2011.

\bibitem{jlxdx2014}
J. Meng, L. Yao, X. Sheng, D. Zhang, and X. Zhu,``Simultaneously
optimizing spatial spectral features based on mutual information for
eeg classification," \emph{IEEE Transactions on Biomedical
Engineering}, vol. 62, no. 1, pp. 227-240, 2014.

\bibitem{xxky2012}
X. J. Zhang, X. M. Zhao, K. He, L. Lu, Y. W. Cao, J. D. Liu, J. K.
Hao, Z. P. Liu, and L. N. Chen, ``Inferring gene regulatory networks
from gene expression data by path consistency algorithm based on
conditional mutual information," \emph{Bioinformatics}, vol. 28, no.
1, pp. 98-104, 2012.

\bibitem{tj1991}
T. M. Cover and J. A. Thomas, Elements of Information Theory.
NewYork: Wiley, 1991.

\bibitem{jth2007}
J. Friedman, T. Hastie, H. Hofling, and R. Tibshirani, ``Pathwise
coordinate optimization". \emph{Annals of Applied Statistics}, no.
1, vol. 2, pp. 302-332, 2007.

\bibitem{matsuo2003}
T. Matsuo, K. Kuriyama, Y. Miyazaki, S. Yoshida, M. Tomonaga,
et al, ¡°The percentage of myeloperoxidase-positive blast cells is a
strong independent prognostic factor in acute myeloid leukemia,
even in the patients with normal karyotype¡±. Leukemia, vol. 17, no.
8, pp. 1538-1543, 2003.

\end{thebibliography}
\end{document}